\def\year{2021}\relax
\documentclass[letterpaper]{article} % DO NOT CHANGE THIS
\usepackage{aaai21}  % DO NOT CHANGE THIS
\usepackage{times}  % DO NOT CHANGE THIS
\usepackage{helvet} % DO NOT CHANGE THIS
\usepackage{courier}  % DO NOT CHANGE THIS
\usepackage[hyphens]{url}  % DO NOT CHANGE THIS
\usepackage{graphicx} % DO NOT CHANGE THIS
\usepackage[switch]{lineno}

\usepackage{hyperref}
\usepackage{multirow}
\usepackage{amsmath}
\usepackage[linesnumbered,ruled,vlined]{algorithm2e}
\SetKwInput{KwInput}{Input}                % Set the Input
\SetKwInput{KwOutput}{Output}              % set the Output
\newcommand{\etal}{\textit{et al}.}

\usepackage{natbib}  % DO NOT CHANGE THIS AND DO NOT ADD ANY OPTIONS TO IT
\usepackage{caption} % DO NOT CHANGE THIS AND DO NOT ADD ANY OPTIONS TO IT
\title{Decision-based Universal Adversarial Attack}
\author{Jing Wu, Mingyi Zhou, Shuaicheng Liu, Yipeng Liu, Ce Zhu\\
	\textit{University of Electronic Science and Technology of China}\\
	{\tt\small wujing@std.uestc.edu.cn}
}
\begin{document}
\maketitle

\begin{abstract} 
	A single perturbation can pose the most natural images to be misclassified by classifiers. In black-box setting, current universal adversarial attack methods utilize substitute models to generate the perturbation, then apply the perturbation to the attacked model. However, this transfer often produces inferior results. In this study, we directly work in the black-box setting to generate the universal adversarial perturbation. Besides, we aim to design an adversary generating a single perturbation having texture like stripes based on orthogonal matrix, as the top convolutional layers are sensitive to stripes. To this end, we propose an efficient Decision-based Universal Attack (DUAttack). With few data, the proposed adversary computes the perturbation based solely on the final inferred labels, but good transferability has been realized not only across models but also span different vision tasks. The effectiveness of DUAttack is validated through comparisons with other state-of-the-art attacks. The efficiency of DUAttack is also demonstrated on real world settings including the Microsoft Azure. In addition, several representative defense methods are struggling with DUAttack, indicating the practicability of the proposed method\footnote{The code is available at \href{https://github.com/zhoumingyi/DUAttack}{GitHub}}.
\end{abstract}

\section{Introduction}

Many challenging practical tasks such as image recognition~\cite{szegedy2015going,he2016deep}, object detection~\cite{ren2015faster}, semantic segmentation~\cite{sun2019high} and so forth, are increasingly relied on Deep Neural Networks (DNNs). However, recent studies have found that DNNs are vulnerable to adversarial examples, images with subtle changes~\cite{szegedy2013intriguing,goodfellow2014explaining,papernot2017practical}. Moreover, these constructed adversarial examples generalize well across networks~\cite{papernot2016transferability}. The usage of these adversarial attacks in real-world applications is problematic, and so motivates the studies on designing robust models resistant to adversarial attacks~\cite{buckman2018thermometer,guo2018countering,akhtar2018defense}.

\begin{figure}[t]
	\centering
	\includegraphics[width=0.42\textwidth]{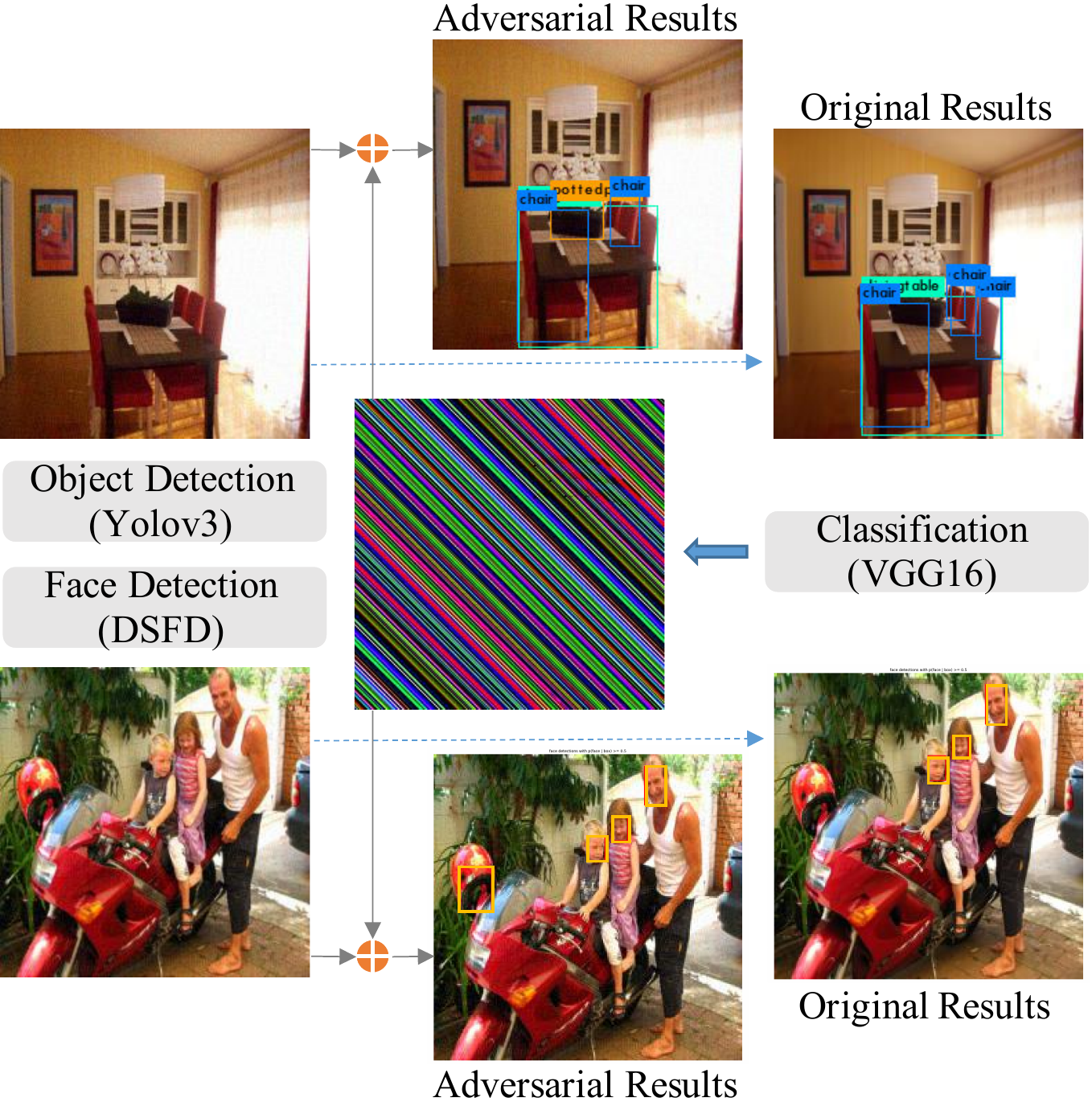}
	\caption{An example of the universal adversarial attack. The universal attack first computes a single perturbation for a classification task, then the perturbation can be added to different natural images, yielding adversarial images. The universal perturbation can be directly applied to different systems such as object detection and face detection. Here, our adversary only has access to the final labels returned by the classifier, but the adversarial perturbation generalizes well across data, models, and tasks. Pixel values of the perturbation are scaled for visibility.}
	\label{fig_intro}
\end{figure}
Moosavi-Dezfooli \etal~\cite{moosavi2017universal} first found a single perturbation that fools DNNs over multiple natural images. They constructed the image-agnostic perturbations by iteratively computing the directions in the decision boundary of the classifier.
Further, Khrulkov and Oseledets~\cite{khrulkov2018art} crafted the universal adversarial perturbations through finding the singular vectors of the Jacobian matrices of hidden layers in a network, while Hayes and Danezis~\cite{hayes2018learning} introduced a generative network for learning the universal adversarial perturbations to fool an attacked model. These works also show that the universal adversarial perturbations constructed by a model can perform well on other models.

The application potential of the universal adversarial attack is higher than the individual attack because the former method only needs to craft one perturbation for most natural images at a time, while the latter one needs to generate the corresponding adversarial perturbation for every natural image. As illustrated in Fig.~\ref{fig_intro}, the universal adversarial attack can attack different systems using one perturbation which is constructed for VGG16~\cite{simonyan2014very}, the perturbation can be added to different natural images. A practical attack method should require as little as possible knowledge of the attacked models, which include training data, models weights, output probabilities and hard labels~\cite{athalye2018obfuscated}. However, existing universal adversarial attack methods need the internal information of the attacked models. For achieving black-box attack, this perturbation is applied to different images and models based on the transferability of the adversarial perturbation. Such a transfer will reduce the performance of the universal perturbation. Therefore, the goal of this study is to generate the perturbation in black-box setting. Besides, these universal adversarial attack methods compute the perturbation having random noise. But literature show that the top convlolutional layers are sensitive to stripes~\cite{krizhevsky2012imagenet}. Athalye \etal~\cite{athalye2018synthesizing} generated adversarial examples with a few color stripe, and got strong misleading performance. Therefore, we aim to use such sensitivity to craft a single perturbation having stripes texture like the one in Fig~\ref{fig_intro}. This mechanism will improve the performance of universal perturbation in black-box setting, and also can be applied to different images, models and tasks.

In this work, we propose a Decision-based Universal Attack (DUAttack) that has no access to the architecture of models and class probabilities and scores. The proposed attacker only has access to the top-1 inferred labels returned by models, which are used to guide the direction for computing the perturbation. However, with such a few access, the experiments show that our adversary performs better than the leading competitors, UAP (Universal Adversarial Perturbations)~\cite{moosavi2017universal} and SingluarFool~\cite{khrulkov2018art}, which has knowledge of the hidden layers of the attacked model. Moreover, a single adversarial perturbation is generated for all natural images, but experiments show that our method have higher transferability than methods that construct the corresponding adversarial examples for every natural image, such as SimBA (Simple Black-box Attack)~\cite{guo2019simple}. Specifically, with a mere few training images, to craft perturbation having stripes texture, we iteratively sampling the predefined orthogonal matrix as the update points, and altering the update vector based on its previous directions. Experiments show that the proposed image-agnostic adversarial attack not only generalizes well across models but also performs well across different tasks. To recap, we propose an algorithm for crafting image-agnostic adversarial perturbation directly under the black-box setting. The main contributions are as follows:
\begin{itemize}
	\item Our adversary only has access to the top-1 labels returned by models, which is the minimum access compared with previous typical black-box attacks.
	\item Our adversary is the first decision-based universal adversarial attack. It generates a single perturbation with stripes texture for all natural images but can outperform the competitive opponent that constructs adversarial examples for every image.
% 	\item Our adversary generated a single perturbation having stripes texture, which are validated more effective than perturbation having random noise for universal attack in our experiments. 
	\item Our experiments not only show that the proposed method generalizes well across images and models, but also performs well across tasks on real-word settings.
\end{itemize}

\section{Related work}
Szegedy \etal~\cite{szegedy2013intriguing} first found the vulnerability of deep neural networks to visually imperceptible perturbations, despite their high accuracy in the context of image classifications. Such adversarial examples can be easily misclassified with high confidence. Even worse, Papernot \etal~\cite{papernot2016transferability} showed that the same adversarial examples generated from one model can fool multiple models. These interesting properties motivate researchers to study attacks and defenses for deep neural networks~\cite{akhtar2018threat}. In general, adversarial attacks can be categorized into individual attacks that construct image-specific perturbations and universal attacks which craft an image-agnostic one.

\subsection{Individual Attacks}
Individual attacks construct adversarial perturbations for every natural image. Among these attacks, gradient-based attacks have access to the architecture of models. Goodfellow \etal~\cite{goodfellow2014explaining} proposed FGSM (Fast Gradient Sign Method), a method computing the gradients with respect to the input of the model. Kurakin \etal~\cite{kurakin2016adversarial} proposed BIM (Basic Iterative Method) which is an iterative version of FGSM. FGSM takes a large step to update the perturbation, instead, BIM iteratively takes small steps. FGSM and BIM aim to find the adversarial examples increasing the loss of the classifier, while DeepFool~\cite{moosavi2016deepfool} constructed the perturbation with minimal norm according to the decision boundaries of the classifier. Carlini and Wagner~\cite{carlini2017towards} also intended to find quasi-imperceptible perturbations by restricting the $\ell_p(p=0,2,\infty)$ norms of the perturbations. Motivated by the study on the optimization landscape, Madry \etal~\cite{madry2017towards} proposed PGD (Projected Gradient Descent), an iterative method to project the perturbations onto a ball of specified radius. Dong\etal~\cite{dong2017boosting} proposed momentum iterative gradient-based methods which have good transferability. With no any real data, Zhou \etal~\cite{zhou2020dast} learned a substitute model to theft information of the attacked model when they only have the output returned by the attacked model, then generated adversarial examples based on gradient-based attack methods through the substitute model. Their data-independent adversarial perturbation has good transferability.

Different from gradient-based attacks, score-based and decision-based attacks only access the class probabilities or the final class predictions. Chen \etal~\cite{chen2017zoo} proposed ZOO (Zeroth Order Optimization) which instead craft the adversarial examples by directly estimating the gradients of the attacked model with zeroth order stochastic coordinate descent. Ilyas \etal~\cite{ilyas2018black} showed an attack NES that access to the probabilities and an attack NES-LO only access to top k ($k \ge 1$) labels. They estimated the gradients of the attacked classifier using NES~\cite{salimans2017evolution} and then applied PGD for generating adversarial examples. Also inspired by NES, instead of gradients estimation, Li \etal~\cite{li2019nattack} built a probability density distribution over the $\ell_p$ ball centered around an input. They assumed that all imperceptible adversarial examples of this input belong to this ball. Likewise, without estimating the gradients, Guo \etal~\cite{guo2019simple} proposed a simple iterative technique SimBA (Simple Black-box Attack). They defined an orthonormal basis and randomly select one, then added or subtracted the chosen point to the input according to the confidence scores. Having only access to the labels returned by the attacked model, Brendel \etal~\cite{brendel2017decision} proposed Boundary Attack that first finds a perturbation with large norms, and then reduces the perturbation. Their attack almost needs no hyperparameter tuning and performs closing to the gradient-based attacks. Chen \etal~\cite{chen2020rays} proposed the Ray Searching attack (RayS), which improves the hard-label attack effectiveness as well as efficiency. Besides, their RayS attack could also help identify falsely robust models.
%Chen \etal~\cite{chen2019hopskipjumpattack} proposed HopSkipJump, an advanced version of Boundary Attack. They estimated the gradient direction using the perturbations inspired by zeroth-order optimization.

\subsection{Universal Attacks}
Contrary to the individual attacks, universal attacks craft a single perturbation for all benign images. Some of these attacks require samples from the data to construct the universal perturbation~\cite{chaubey2020universal, moosavi2017universal,moosavi2018robustness,khrulkov2018art,hayes2018learning,reddy2018nag,poursaeed2018generative}. Moosavi-Dezfooli \etal~\cite{moosavi2017universal} first introduced the universal adversarial perturbation. They directly crafted the image-agnostic perturbation through DeepFool~\cite{moosavi2016deepfool} and showed some interesting properties such as transferability across models. Khrulkov and Oseledets~\cite{khrulkov2018art} proposed to compute the so-called $(p,q)$-singular vectors of the Jacobian matrices of feature maps in the attacked model for generating the universal perturbation. Hayes and Danezis~\cite{hayes2018learning} designed universal adversarial networks based on a generator to construct the universal adversarial perturbation.

Besides, some researchers study on crafting the data-free universal adversarial perturbation~\cite{mopuri-bmvc-2017,reddy2018ask,mopuri2018generalizable}. Mopuri \etal~\cite{mopuri-bmvc-2017} crafted the data-independent universal adversarial perturbation by maximizing the mean activations at multiple layers of the attacked model. Mopuri \etal~\cite{reddy2018ask} generated class impressions by starting from a random noise and update the noise until the attacked model misclassifies it with high confidence. Then they utilized the class impressions as training data for a generator to construct the universal adversarial perturbation.

Methods above for constructing the universal adversarial perturbation need the information about the attacked model internals. In real-world systems, the adversary has little chance to know about the attacked model internals. So current universal attacks generate perturbation from a substitute model, then apply the perturbation to the attacked models for achieving black-box attacks. However, such a transfer will degrade the performance of the attacker. As such, we explore, in this paper, the construction of universal adversarial perturbation directly in the back-box setting.

\section{DUAttack}
\textbf{Notations and Preliminaries.}
Before introducing the proposed method, we present the notations and preliminaries. A scalar, vector and matrix are written as $a$, $\mathbf{a}$ and $\mathbf{A}$ respectively. Below, we formalize the notions of universal perturbations and present the method for constructing the perturbations directly in the black-box setting. Suppose that $\mathcal{U}$ denotes the distribution of the natural images in $\mathcal{R}^{H \times W \times C}$. Let $\mathcal{F}(\cdot)$ denote a classifier that maps each image $\mathbf{X} \in \mathcal{U}$ to its final estimated label $y \in \mathcal{R}$. We aim to craft a perturbation $\mathbf{V} \in \mathcal{R}^{H \times W \times  C}$ which causes most images sampled from the distribution $\mathcal{U}$ misclassified by the classifier $\mathcal{F}$. The universal perturbation $ \mathbf{V} $ satisfies the following constraints:
\begin{equation}
\label{eq_uap}
\mathop{\mathcal{P}}\limits_{\mathbf{X} \sim \mathcal{U}}(\mathcal{F}(\mathbf{X}) \ne \mathcal{F}(\mathbf{X}+\mathbf{V})) \ge \delta \quad s.t. \left\|\mathbf{V}\right\|_p \le \zeta,
\end{equation}
where $\delta \in (0, 1]$ quantifies the desired fooling ratio, $p \in [1, \infty)$, $\left\|\cdot\right\|_p$ and $\mathcal{P}(\cdot)$ denote the $\ell_p$ norms and the probability respectively, and the attack distance $\zeta \ge 0$ is a constant. In this study, the hyper-parameters of the porposed method DUAttack are the step size $ \epsilon $ and the number of iteration $ T $.

\vspace{+0.5em}
\noindent
\textbf{Scenario.}
This study focuses on synthesizing the universal adversarial perturbation by decision-based attack. In current studies, universal attacks need to use a substitute model to generate the perturbation through gradient optimization methods, then apply the perturbation to the attacked models for achieving the black-box attack. In our study, the proposed method only access the inferred top-1 labels returned by the attacked models. The attacker can generate the universal perturbation for the attacked models directly.

\vspace{+0.5em}
\noindent
\textbf{Algorithm.}
We now introduce the proposed method for crafting the universal perturbations using Algorithm~\ref{alg_dduap}. $\{\mathbf{X}^i\}(i=1,2,...,N)=\{\mathbf{X}^1, \mathbf{X}^2, ..., \mathbf{X}^N\}$ is a set of clean images sampled from the data distribution $\mathcal{U}$, its number is $N$. The direction for updating the perturbation of DUAttack is inspired by SimBA~\cite{guo2019simple}, which is a powerful query-based individual attack method. SimBA utilizes the output probabilities returned by the attacked models, instead, we only have access to the final inferred top-1 labels $\mathbf{y}=\mathcal{F}(\{\mathbf{X}^i\})$ returned by the classifier. Besides, we design the orthogonal matrix and momentum-based optimization for efficiently computing the universal perturbation. We iteratively compute the universal perturbation $\mathbf{V}$ with the Frobenius norm $\zeta$, which can cause most test data different from $\{\mathbf{X}^i\}$ misclassified by the attacked model.

Under the black-box setting, the key idea of DUAttack is that the labels returned by the model can guide the direction for crafting the perturbation. See pseudo-code in Algorithm~\ref{alg_dduap}, in each iteration, 1) at first, we randomly utilize an orthogonal matrix $\mathbf{Q}$ to decide which points should be altered, compute the perturbation with a dynamical step size $\epsilon*\mathbf{Q} + 0.9 * \mathbf{M}$ (0.9 is an empirical value), where $\mathbf{M}$ records the historical changes of the perturbation. 2) Secondly, to make sure the perturbation will always satisfy the constraint $\left\|\mathbf{V}\right\|_F \le \zeta$, we further project the updated perturbation onto a sphere centered at 0 and with a radius $\zeta$. 3) Thirdly, we subtract and add the adversarial examples with the perturbation respectively, compare the final inferred labels $\mathcal{F}(\mathbf{X})$ and $\mathcal{F}(\mathbf{X}-\mathbf{V})$ or $\mathcal{F}(\mathbf{X}+\mathbf{V})$ corresponding with the clean input $\mathbf{X}$ and adversarial input $\mathbf{X}-\mathbf{V}$ or $\mathbf{X}+\mathbf{V}$ respectively, and record the numbers $\mathbf{re}_l$ and $\mathbf{re}_r$ of adversarial examples $\mathbf{X}-\mathbf{V}$ and $\mathbf{X}+\mathbf{V}$ that successfully fool the model respectively. 4) Finally, we choose the larger one where adversarial examples successfully make the model output labels different from their counterparts do, and update the momentum matrix $\mathbf{M}$ with these altered points in this iteration.

\begin{algorithm}[tb]
\SetAlgoLined
	\KwInput{Classifier $\mathcal{F}$, the step size $\epsilon$, the maximum number of iterations $T$, and the desired Frobenius norm of the perturbation $\delta_m$, data points $\{\mathbf{X}^i\}(i=1,2,...,N) \in \mathcal{U}$,  orthogonal matrix $\mathbf{Q} \in \mathcal{R}^{H \times W \times C}$\, and momentum matrix $\mathbf{M} \in \mathcal{R}^{H \times W \times C}$.}
	\KwOutput{Universal adversarial perturbation $\mathbf{V}$}
	Initialization: $\mathbf{V}\Leftarrow \mathbf{0}$, $\mathbf{M}\Leftarrow \mathbf{0}$,  $\mathbf{Q}\Leftarrow \mathbf{E}$, $k\Leftarrow 0$ \;
	\For{$t \leftarrow 1 $ \KwTo $T$}{
			$ random\quad k \in [0,W)$\;
			$ \mathbf{Q} \gets (\mathbf{e}_k, \mathbf{e}_k+1, \mathbf{e}_{W-1}, \mathbf{e}_0, ..., \mathbf{e}_{k-1}) $\;
			$vector\quad \mathbf{re}_l \gets 0, \mathbf{re}_r \gets 0$\;
			$\mathbf{V}^{t}_l \gets \mathbf{V}^{t-1} - (\epsilon \mathbf{Q} +0.9 \mathbf{M})$\;
			$\mathbf{V}^t_l = \delta_m \mathbf{V}^{t-1} / \left\|\mathbf{V}^{t-1}\right\|_F$\;
		    $\mathbf{re}_l  \gets \mathcal{F}(\mathbf{X}) \ne \mathcal{F}(\mathbf{X}-\mathbf{V}^t_l)$\;
			\If{$sum(\mathbf{re}_l) = 0$}{$\mathbf{V}^t = \mathbf{V}^t_l$; \quad break\;}
			$\mathbf{V}^{t}_r \gets \mathbf{V}^{t-1} + (\epsilon \mathbf{Q} +0.9 \mathbf{M})$\;
			$\mathbf{V}^t_r = \delta_m \mathbf{V}^{t-1} / \left\|\mathbf{V}^{t-1}\right\|_F$\;
			$\mathbf{re}_r \gets \mathcal{F}(\mathbf{X}) \neq \mathcal{F}(\mathbf{X}+\mathbf{V}^t_r)$\;
			\If{$sum(\mathbf{re}_r) = 0$}{$\mathbf{V}^t = \mathbf{V}^t_r$;\quad break\;}
			\eIf{$sum(\mathbf{re}_r) < sum(\mathbf{re}_l)$}{
				$\mathbf{M} \gets \mathbf{M} + \epsilon \mathbf{Q};\quad \mathbf{V}^t = \mathbf{V}^t_l$\;
			}{
			$\mathbf{M} \gets \mathbf{M} - \epsilon \mathbf{Q};\quad \mathbf{V}^t = \mathbf{V}^t_r$\;}
			$t \gets t + 1$ \;
		}
	\Return $\mathbf{V^t - X}.$
\caption{Generating universal peturbation through DUAttack}
\label{alg_dduap}
\end{algorithm}

The proposed algorithm is simple yet practical. It can convert to the targeted attack easily. The only hyper-parameters of the proposed algorithm are the step size $\epsilon$ and the iteration numbers $T$. Besides, the desired $\ell_F$ norm of the perturbation and the number of data points $N$ used to construct the universal adversarial perturbation are set by the need of the attacker itself. The details about the constituents of the proposed method are explained below.

\vspace{+0.5em}
\noindent
\textbf{Orthogonal matrix.}
In the black-box setting, we cannot access the information such as gradients of the attacked models. The conventional methods usually estimate the gradients based on optimization which often needs lots of queries, instead, to craft a single perturbation having stripes texture which let the top convolutional layers more sensitive, we choose a orthogonal unit matrix to decide which pixels to alter for crafting the universal adversarial perturbation.

The size of orthogonal unit matrix is the same as the inputs, and each column vector of the matrix is orthogonal. For example, suppose that the resolution of one image is $H \times W$, the size of the matrix will also be $H \times W$. We have $H \times 1$ vector $\mathbf{e}_i=(0 \ 0 \cdots 1 \cdots 0)^T(i=0,1,2,\dots,W-1)$ where the entries in the $i$-th position of the vector is $1$ and other positions have $0$. At each iteration, we randomly select a column vector and then move the whole matrix a few steps to the left or right so that the selected vector can be the first column vector. For instance, without loss of generality, suppose that $H=W$, the initial orthogonal unit matrix is $(\mathbf{e}_0 \ \mathbf{e}_1 \cdots \mathbf{e}_{W-2} \ \mathbf{e}_{W-1})$. In current iteration $t=1$, we choose the $2$-th column vector $\mathbf{e}_2$, then the matrix will be $(\mathbf{e}_2 \ \mathbf{e}_3 \cdots \mathbf{e}_{W-1} \ \mathbf{e}_1 \ \mathbf{e}_0)$.

In the proposed method, in each iteration, we compute one channel of the perturbation with the selected orthogonal matrix. Thus, we alter $W$ pixels of the perturbation one time. Due to the randomness, every pixel has the same chance to be altered after several iterations.

\vspace{+0.5em}
\noindent
\textbf{Update strategy.}
Previously, the selected pixels are all changed with the same steps. In the proposed method, this may cause the problem that some pixels will offset each other due to the subtraction or plus in different iterations. Therefore, inspired by the optimization method with momentum~\cite{ruder2016overview} in deep learning, we apply the same idea here. The parameter of momentum is 0.9 in our method. In general, the altered pixels have dynamical step size, it can restrain such offset and reduce the number of queries. A comparison between different update strategies including the random sampling and the  efficient variant of covariance matrix adaptation evolution strategy (CMAES)~\cite{dong2019efficient} can be seen in Section~\ref{subsec_us}.

\vspace{+0.5em}
\noindent
\textbf{Projection.}
Notice that our final goal is to find a single perturbation for all natural images, the Frobenius norm of the perturbation can be large under no constraints.

For Algorithm~\ref{alg_dduap}, images in a mini-batch update the perturbation with $\epsilon$. Given that we use $N$ images to construct the universal adversarial perturbation and a mini-batch has $n$ images, the perturbation will be updated by $q=\lceil N/n \rceil$ times. The size of the perturbation $\mathbf{V}$ is $H \times W \times C$, the maximum number of iterations is $T$. The size of the unit matrix $\mathbf{Q}$ is $H \times W$. We assume that the perturbation is updated in the same direction every time and $H=W=m$, we do not consider the momentum term here, therefore the Frobenius norm of the final perturbation can be
\begin{equation}
\label{eq_l2}
\delta = \left\|\mathbf{V}\right\|_F = \left\|\sum_{i=1}^q\sum_{j=1}^T\epsilon \mathbf{Q}\right\|_F.
\end{equation}
So the norm of the perturbation will be
\begin{equation}
\label{eq_l21}
\delta = \left\|\sum_{i=1}^q\sum_{j=1}^T\epsilon \mathbf{Q}\right\|_F \le \epsilon\sqrt{qTm}.
\end{equation}

Notice that $m$ is equal to the size of the images, the norm of the perturbation and the number of the iteration are positively correlated. Under this circumstance, the adversary cannot attack imperceptibly. Thus we consider to constraint the norm of the perturbation less than $\zeta$ through projection
\begin{equation}
\label{eq_nor}
\mathbf{V} \gets arg\mathop{min}\limits_{\mathbf{V}'}\left\|\mathbf{V}-\mathbf{V}'\right\|_F \quad s.t.\left\|\mathbf{V}'\right\|_F \le \zeta,
\end{equation}
which can ensure the Frobenius norm of the perturbation to satisfy the constraint $\left\|\mathbf{V}'\right\|_F \le \zeta$ in each iteration.

DUAttack constructs the universal adversarial perturbation using Algorithm~\ref{alg_dduap}, then adds unseen natural images with the perturbation to fool models. The proposed method only needs queries in the synthesis phase, the universal perturbation can be applied to different images, models and tasks on real-world settings.  

\section{Experimental Evaluation}
In this section, we conduct various experiments on MNIST~\cite{lecun1998gradient}, CIFAR-10~\cite{krizhevsky2009learning}, ImageNet~\cite{deng2009imagenet} and PASCAL VOC2007~\cite{pasca_voc_2007} to investigate the effectiveness of DUAttack. 
\begin{itemize}
	\item To evaluate the effectiveness and transferability of the adversarial examples, we perform the ablation study with VGG16~\cite{simonyan2014very} and ResNet50~\cite{he2016deep}, we compare DUAttack with other approaches, including two universal attacks UAP~\cite{moosavi2017universal} and SingularFool~\cite{khrulkov2018art}, and three individual attacks FGSM~\cite{goodfellow2014explaining}, PGD~\cite{madry2017towards} and SimBA~\cite{guo2019simple}. 
	\item To investigate the generality property of cross models, we compare the fooling rates on different attacked models using different models to craft the universal perturbation. Furthermore, to investigate the universality of cross tasks, we apply the universal perturbation generated from the classification task to the detection system Yolov3~\cite{redmon2018yolov3}, DSFD~\cite{li2018dsfd}. 
	\item Besides, we analyze the effect of the number of training set for constructing the universal perturbation and compare the performance of the perturbation having stripes and random noise.  We show the performance of DUAttack against several representative defense methods~\cite{guo2018countering,xu2017feature,das2017keeping}. 
	\item Finally, to validate the efficiency of the proposed attack method DUAttack on real-world settings, we compare performance of DUAttack and other attack methods when attacking the online model from Microsoft Azure.
\end{itemize}

\subsection{Experiment Setup}
\textbf{Models.}
In universal and individual attacks comparison, we use VGG16 for all methods to generate the universal adversarial perturbation, which is used to attack ResNet50. In CIFAR-10 experiments, we train models using training images from CIFAR-10. The whole networks are trained with stochastic gradient descent (SGD). All models are trained with an initial learning rate of 0.05, a weight decay of 0.5 after every 30 epochs with a momentum of 0.9. In ImageNet experiments, models are pre-trained on ImageNet. For attack across models, we use VGG16, VGG19~\cite{simonyan2014very}, ResNet50, ResNet101~\cite{he2016deep} and GoogLeNet~\cite{szegedy2015going} pre-trained on Imagenet. For attack across tasks, the attacked models Yolov3 is pre-trained on COCO~\cite{lin2014microsoft}, DSFD is pre-trained on WIDER FACE~\cite{yang2016wider}.

\vspace{+0.5em}
\noindent
\textbf{Datasets.}
The proposed attack method, UAP and SingularFool randomly sample $N(N=500)$ training images from CIFAR-10 or from ImageNet (with different labels) to generate universal adversarial perturbation on VGG16. 10,000 test images from CIFAR-10 or 10,000 validation images (not be included in training) from ImageNet are used for evaluation.
For attack across task, 500 training images from PASCAL VOC2007 are randomly sampled to generate a universal adversarial perturbation. This perturbation is applied to Yolov3 and DSFD on test images of PASCAL VOC2007.

\vspace{+0.5em}
\noindent
\textbf{Parameters.}
For SimBA, UAP and SimgularFool, we use their default settings except for the attack distance controlled parameters. In most experiments, for DUAttack, the step size $\epsilon=0.2$ and the number of iteration $T=1000$.

\vspace{+0.5em}
\noindent
\textbf{Metric.}
The Frobenius norm of the perturbations is the distance between the natural images and the perturbation images, namely $\left\|\mathbf{X}_{adv} - \mathbf{X}\right\|_F$. We use the average distance over images as the metric. Attack success rate under specific distance evaluates the performance of the attack. For the untargeted and targeted attack, we compute the fooling rate $ASR_{untargeted}$ and $ASR_{targeted} $ upon examples that already have been classified correctly by the models and those have not been classified to the targeted label respectively, namely
\begin{equation}
\begin{aligned}
	\label{equ_metrix}
	&ASR_{untargeted} = (\mathcal{F}(\mathbf{X+V}) \ne y) / (\mathcal{F}(\mathbf{X}) = y), \\
	&ASR_{targeted} = (\mathcal{F}(\mathbf{X+V}) = y_t) / (\mathcal{F}(\mathbf{X}) \ne y_t).
\end{aligned}
\end{equation}
For comparsion equally, we show different results with different distance. Besides, the proposed method needs no query in the evaluation phase, and for universal attack comparison, our method is the only one which conducts in the black-box setting, so we do not compare the query.

\begin{table*}[t]
	\begin{center}
		\small
		\caption{Untargeted fooling rates (\%) on universal attack comparison. The universal adversarial perturbation is computed for VGG16 using only 500 images. The numbers in ($\cdot$) denote the average $\ell_F$ perturbation distance per image. Note that when attack VGG16, UAP~\cite{moosavi2017universal} and SingularFool~\cite{khrulkov2018art} are white-box attacks. The universal perturbation crafted on VGG16 then is transfered to ResNet50.}
% Higher fooling rate with smaller distance is better.
		\label{tab_uni}
		\begin{tabular}{lcccccc} 
			\hline
			\multirow{2}{*}{Method} & &\multicolumn{2}{c}{CIFAR-10} & &\multicolumn{2}{c}{ImageNet}\\
			\cline{3-4} \cline{6-7} &
%			&\multicolumn{1}{c}{Queries}
			&\multicolumn{1}{c}{VGG16}&\multicolumn{1}{c}{ResNet50}  &
%			&\multicolumn{1}{c}{Queries}
			&\multicolumn{1}{c}{VGG16}&\multicolumn{1}{c}{ResNet50}\\
			\hline
			%UAP~\cite{moosavi2017universal}  & 
			UAP  & 
%			&\multicolumn{1}{c}{1933}
			&\multicolumn{1}{c}{89.94(15.33)} &\multicolumn{1}{c}{89.61(15.35)}  &
%			&\multicolumn{1}{c}{19470}
			&\multicolumn{1}{c}{57.78(3933.42)}  &\multicolumn{1}{c}{28.25(3928.86)} \\
			%SingularFool~\cite{khrulkov2018art}  & 
			SingularFool& 
%			&\multicolumn{1}{c}{10501}
			&\multicolumn{1}{c}{88.69(13.98)} &\multicolumn{1}{c}{88.61(13.99)}  &
%			&\multicolumn{1}{c}{10501}
			&\multicolumn{1}{c}{56.43(4268.64)} &\multicolumn{1}{c}{30.95(4266.38)} \\		
			DUAttack    &
%			&\multicolumn{1}{c}{2000}
			&\multicolumn{1}{c}{\textbf{91.97(13.23)}} &\multicolumn{1}{c}{\textbf{90.35(13.27)}}  &
%			&\multicolumn{1}{c}{2000}
			&\multicolumn{1}{c}{\textbf{77.36(3911.77)}} &\multicolumn{1}{c}{\textbf{43.14(3912.04)}} \\
			\hline
		\end{tabular}
	\end{center}
\end{table*}
\begin{table*}[t]
	\begin{center}
		\small
		\caption{Untargeted fooling rates (\%) on individual attack comparison. The numbers in ($\cdot$) denote the average $\ell_F$ perturbation distance per image. Note that FGSM~\cite{goodfellow2014explaining}, PGD~\cite{madry2017towards} are white-box attacks when attack VGG16, SimBA~\cite{guo2019simple} is individual score-based attack and DUAttack only use 500 images to craft the universal perturbation. The adversarial examples crafted on VGG16 then is transfered to ResNet50. $1^{st} / 2^{nd}$ best in \textbf{bold}/\textit{italic}. As a desicion-based universal adversarial adversary, DUAttack has good transferability.}.
		\label{tab_ind}
		\begin{tabular}{lcccccc}
			\hline
			\multirow{2}{*}{Method} & &\multicolumn{2}{c}{CIFAR-10} & &\multicolumn{2}{c}{ImageNet}\\
			\cline{3-4} \cline{6-7} &
			&\multicolumn{1}{c}{VGG16} &\multicolumn{1}{c}{ResNet50} &
			&\multicolumn{1}{c}{VGG16} &\multicolumn{1}{c}{ResNet50}\\
			\hline
			%FGSM~\cite{goodfellow2014explaining} & 
			FGSM & 
			&\multicolumn{1}{c}{89.77(15.33)}
			&\multicolumn{1}{c}{49.28(3.28)} &
			&\multicolumn{1}{c}{\textit{94.75(1115.26)}}
			&\multicolumn{1}{c}{\textit{55.60(1115.43)}}\\
			%PGD~\cite{madry2017towards}	& 
			PGD & 
			&\multicolumn{1}{c}{\textit{93.75(3.210)}}
			&\multicolumn{1}{c}{\textbf{72.38(2.60)}}
			&&\multicolumn{1}{c}{\textbf{96.10(1008.47)}}  &\multicolumn{1}{c}{\textbf{39.74(1008.44)}} \\
			%SimBA~\cite{guo2019simple} & 
			SimBA & 
			&\multicolumn{1}{c}{\textbf{98.44(1.940)} }
			&\multicolumn{1}{c}{27.70(3.14)}
			&&\multicolumn{1}{c}{94.70(1187.92)} &\multicolumn{1}{c}{2.74(1187.92)} \\	
			DUAttack    &
			&\multicolumn{1}{c}{91.97(13.23) }
			&\multicolumn{1}{c}{\textit{53.26(3.28)}}
			&&\multicolumn{1}{c}{77.36(3911.77)}
			&\multicolumn{1}{c}{7.99(1122.93)}\\
			\hline
		\end{tabular}
	\end{center}
\end{table*}
\subsection{Attack methods comparison}
\textbf{Universal attack comparison}
We compare the performance of the proposed attack method DUAttack with the universal adversarial attack methods UAP and SingularFool in the same setting. We use $N=500$ clean images randomly sampled from the training images in CIFAR-10 or sampled from the validation images in ImageNet to compute the universal adversarial perturbation through VGG16. Then we evaluate 10,000 images sampled from the testing images in CIFAR-10 or the other validation images in ImageNet with the perturbation on ResNet50.

As shown in Table~\ref{tab_uni}, the proposed attacked method has better performance than UAP and SIngularFool on CIFAR-10. Notice that when attacking VGG16, UAP and SingularFool are in the white-box setting. On ImageNet, when the attack distances are similar, the fooling rate of DUAttack on VGG16 is about 20\% higher than that of UAP and SingularFool. When transfered to ResNet50, the fooling rate of the universal perturbation crafted by DUAttack is more than 10\% than that of UAP and SingularFool. These results show that the adversarial examples generated by the proposed method DUAttack have higher transferability than those constructed by UAP and SingularFool.

\vspace{+0.5em}
\noindent
\textbf{Individual attack comparison}
Besides, we also compare DUAttack with individual attack methods on CIFAR-10 and ImageNet. We use Advertorch~\cite{ding2019advertorch} to perform FGSM and PGD. As shown in Table~\ref{tab_ind}, generally (ReNet50 on CIFAR-10, ResNet50 on ImageNet), our proposed adversary has higher transferability than the state-of-the-art individual score-based attacker SimBA. DUAttack even has better performance than the individual white-box attackers FGSM when attacking VGG16 on CIFAR-10. Note that DUAttack only has access to the final inferred labels returned by the models and aims to find a single perturbation for every natural image, while the individual white-box attackers have knowledge about the hidden layers of the models and meant to find different perturbations for different images. Compare with the individual black-box attacker, our method only produces the universal perturbation once on the whole dataset, the individual black-box attacker needs to generate the corresponding perturbation on every image.

\begin{figure}[thb]
	\centering
	\includegraphics[width=0.42\textwidth]{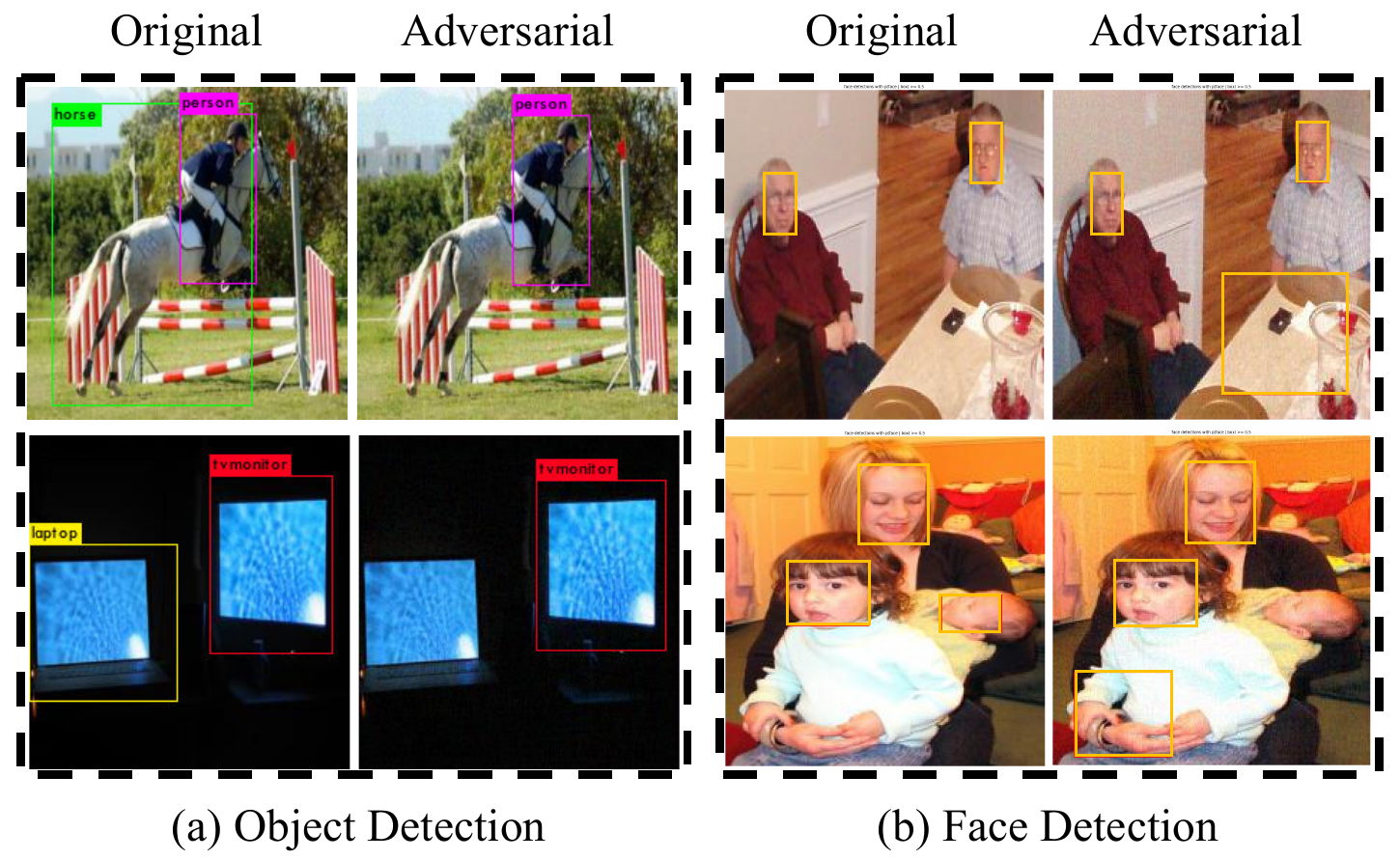}
	\caption{Detection results on Yolov3~\cite{redmon2018yolov3} and DSFD~\cite{li2018dsfd}. (a) Results of clean images and adversarial images detected by Yolov3 are in the first column and second column respectively. (b) Results of clean images and adversarial images detected by DSFD are in the first column and second column respectively. The universal perturbation learned in the classification task can also be applied to the detection task with a good attack performance. See supplementary material for additional samples.}
	\label{fig_task}
\end{figure}
\subsection{Universality across models}
\begin{table*}[tb]
	\begin{center}
		\small
		\caption{Untargeted fooling rates (\%) of the perturbations crafted for one model on other models using only 500 images. The rows indicate the models for which the universal perturbations are computed, and the columns indicate the attacked models for which the fooling rates are reported. The second column shows the top-1 accuracy (\%) of the models on ImageNet. The perturbation computed by DUAttack generalizes well across models.} 
		\label{tab_model}
		\begin{tabular}{l|c|ccccc|} 
		\cline{2-7}
		&Accuracy  &VGG16   &VGG19    &ResNet50    &ResNet101   &GoogLeNet \\
		\hline
		\multicolumn{1}{|c|}{VGG16} 	&72.11	&\textbf{77.36} &68.45 	&43.14 	&38.40 &36.72\\
		\multicolumn{1}{|c|}{VGG19} 	&73.00	&\textbf{69.69} &65.10  &40.50 	&36.31 &38.89\\
		\multicolumn{1}{|c|}{ResNet50} 	&76.65	&\textbf{63.16} &55.85 	&40.93  &32.87 &32.33\\
		\multicolumn{1}{|c|}{ResNet101} &77.61	&\textbf{62.74} &56.64 	&37.93 	&32.82 &32.33\\
		\multicolumn{1}{|c|}{GoogLeNet} &70.23	&\textbf{62.85} &57.28 	&35.63  &30.00 &34.63\\
		\hline
		\end{tabular}
	\end{center}
\end{table*}
We have shown that the perturbation crafted by DUAttack generalize well across images, now we investigate its universality across models. We choose 5 popular models, construct the universal perturbation for each network, and report the fooling rates on other models with this perturbation on ImageNet. The number of iterations, the step size for the proposed attack method are 1000, 0.2 respectively. The desired Frobenius norm of the perturbation is about 4000  and the number of images for computing the perturbation is 500.

As shown in Table~\ref{tab_model}, the rows are the models constructing the perturbation, and the columns are the models that being attacked with the perturbation. Table~\ref{tab_model} indicates that the perturbation constructed by a network can generalize very well across other models, all perturbations computed by a network have a fooling rate over 30\% on other models on ImageNet. Fooling rates in Table~\ref{tab_model} indicate that the attack performance of DUAttack is related to the capacity and the classification accuracy of the models. VGG series are more likely to be cheated. Despite the model VGG series, models with high classification accuracy have weak robustness, which indicates a trade-off between the accuracy and robustness of the models. Thorough discussions can be seen in the articles~\cite{zhang2019theoretically, raghunathan2020understanding}.

Results in Table~\ref{tab_model} shows the great application potential of DUAttack, as the universal perturbation generalize well across systems with only access to the top-1 inferred labels returned by the model.
\subsection{Universality across tasks}
\begin{figure}[t]
	\centering
	\includegraphics[width=0.45\textwidth]{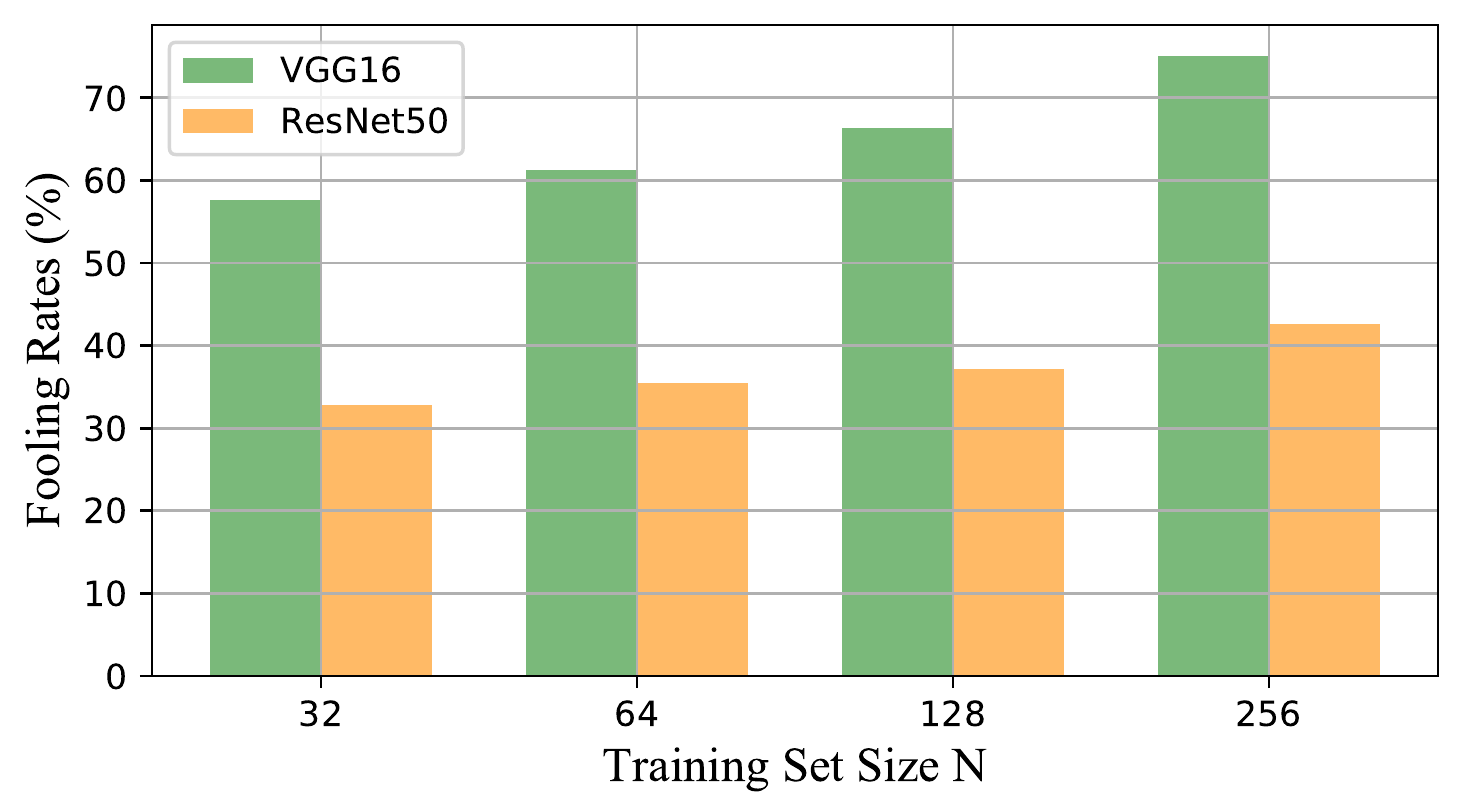}
	\caption{Fooling rates on the 10,000 validation set from ImageNet versus the training set size of $N$ for computing the universal adversarial perturbation.}
	\label{fig_trainsize}
\end{figure}
\begin{figure}[t]
	\centering
	\includegraphics[width=0.45\textwidth]{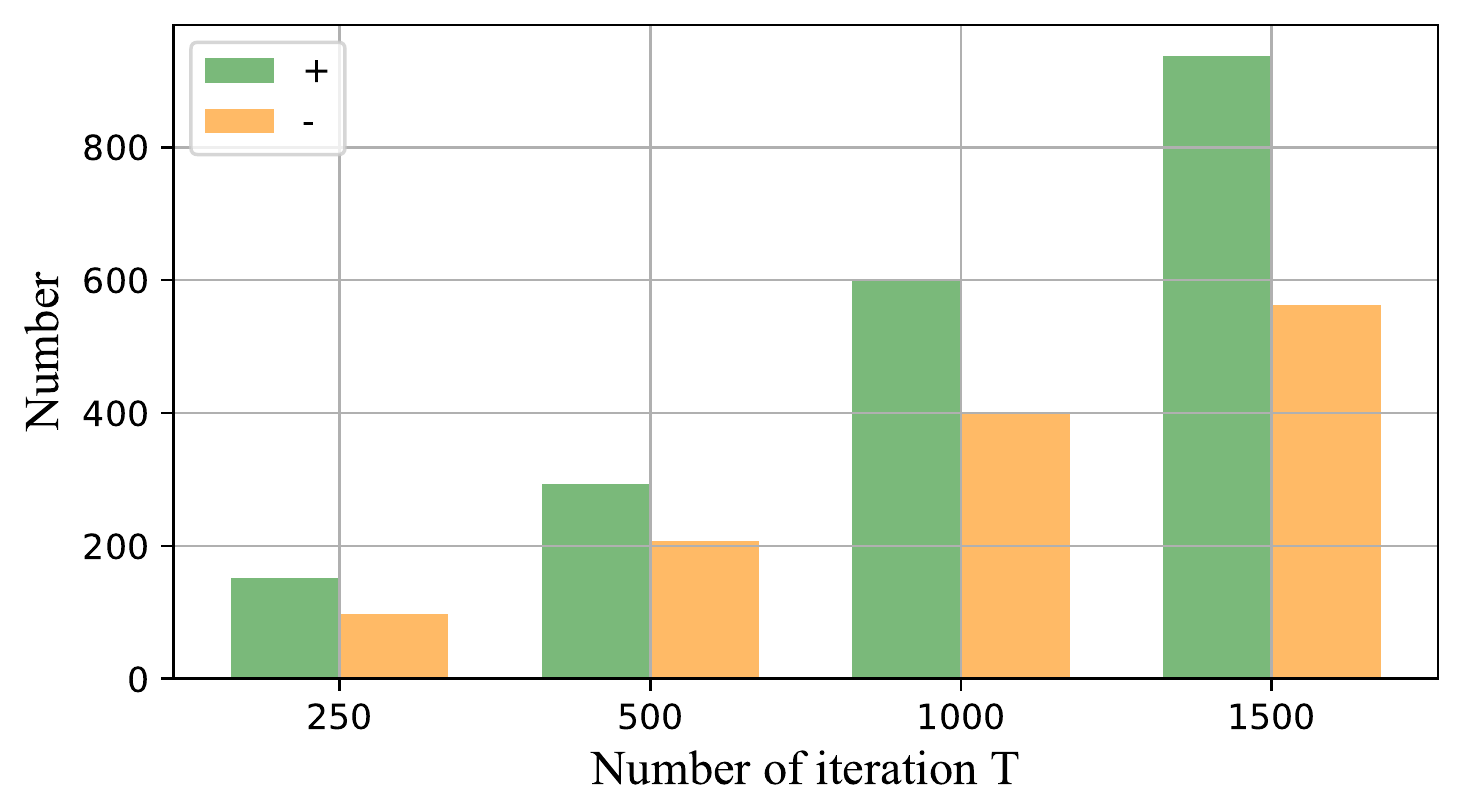}
	\caption{Distribution of plus and minus operations during the updating process when the proposed method DUAttack has different number of iterations.}
	\label{fig_plusminus}
\end{figure}
We have shown that our DUAttack is doubly-universal as the perturbation crafted for a model generalizes well across data and models. Now we further estimate the performance of the universal perturbation on different tasks.

Specifically, 1) we randomly sample 500 images from training images in PASCAL VOC, utilize these images to generate a universal perturbation for VGG16 under the multi-label classification task. 2) We choose Yolov3 trained on COCO, DSFD trained on WIDER FACE to be the attacked models. 3) Then we apply the universal adversarial perturbation to randomly 64 test images. 4) Finally, we utilize these adversarial images to attack Yolov3 and DSFD.

Among 64 test images, 35 images are detected incorrectly by Yolov3. Some detected results are shown in Fig.~\ref{fig_task} (a). For Yolov3, some objects are missed detection. Nevertheless, the horse in the first row of (a) and the TV monitor in the second row of (a) are obvious in the adversarial images, however, Yolov3 still cannot detect them. Then, from the test images, we pick out images with faces of people. We use DSFD to detect these images. some results are shown in Fig.~\ref{fig_task} (b). In the first row of (b), DSFD has false detection on the table. In the second row of (b), DSFD cannot detect the face of the baby and has false detection on the hands of the girl. More examples are included in the supplementary material.

Fig.~\ref{fig_task} shows that adversarial examples generated through the proposed method have high transferability. The universal adversarial perturbation is trained on multi-label classification task with VGG16, after being applied to test images from PASCAL VOC, even can fool Yolov3 trained on COCO and DSFD trained on WIDER FACE.

\subsection{Effect of training set size}
Experiments above have shown that the proposed method DUAttack is triple-universal, as the perturbation can generalize well across data, models, and tasks. We now investigate the performance of DUAttack with different sizes of the training set. The step size $ \epsilon $ for DUAttack is 0.2, and the number of iteration $ T $ is 1000. We computed the universal adversarial perturbation $ \mathbf{V} $ for VGG16 using a set of $N (N=32, 64, 128, 256)$ images with different labels randomly sampled from the validation images of ImageNet, and then add other 10,000 benign images from ImageNet validation set with the universal adversarial perturbation $ \mathbf{V} $  to generate adversarial images. Then we use these adversarial images as the input of VGG16 and ResNet50.

As illustrated in Fig.~\ref{fig_trainsize}, the fooling rate can be higher if the training set size is larger. We hypothesize that even when using only a set of 32 images to compute the universal adversarial perturbation $ \mathbf{V} $, the proposed attacke method DUAttack can fool more than 50\% of the 10,000 images with less than $ \ell_F = 4000 $ attack distance per image, which is very small for imagenet having images ranging from 0 to 255. In addition, this perturbation $ \mathbf{V} $ can fool more than 30\% of the 10,000 images when transfer to ResNet50. Note that the proposed attacked method only have access to the final inferred top-1 labels from the attacked models, the set of images for computing the universal adversarial perturbation $ \mathbf{V} $ even cannot include all labels on ImageNet (the number of classes is 1000), these results are significant.

\subsection{Analysis of update strategy}
\label{subsec_us}
\begin{table}[tb]
	\begin{center}
		\small
		\caption{Fooling rates (\%) on MNIST when DUAttack using different update strategies. DUAttack\_rand, DUAttack\_CMAES and DUAttack represent the proposed method use random sampling, CMAES~\cite{dong2019efficient} and the proposed orthogonal matrix with momentum respectively.}
		\label{tab_upd}
		\begin{tabular}{l|cc} 
			\hline
			Update Startegy  &Fooling rates (\%)  &Distance \\
			\hline
			DUAttack\_rand   &44.88   &5.09 \\
			DUAttack\_CMAES  &68.36    &4.79 \\
			DUAttack   &\textbf{83.08}  &4.73 \\
			\hline
		\end{tabular}
	\end{center}
\end{table}
For comparing  the effectiveness of the proposed method with different update strategies, we apply random sampling and the covariance matrix adaptation evolution strategy (CMAES)~\cite{dong2019efficient} , as the update strategy of DUAttack. We randomly sample 500 images for computing the universal adversarial perturbation $ \mathbf{V} $ to attack the online model from Microsoft Azure. The step size is 0.02, and the number of iterations is 500.

Results are shown in Table~\ref{tab_upd}. DUAttack\_rand, DUAttack\_CMAES and DUAttack represent the proposed attack method DUAttack using the random sampling strategy, the covariance matrix adaptation evolution strategy and the porposed orthogonal matrix with momentum strategy as the update strategy respectively. As shown in Table~\ref{tab_upd}, the prposed method DUAttack using the orthogonal matrix with momentum as the update strategy can have the best performance. When the attack distance is similiar, DUAttack using the orthogonal matrix with momentum can fool more than 80\% of the 10,000 images, while DUAttack using the random sampling strategy and the covariance matrix adaptation evolution strategy just can fool about 45\% and 68\% of the 10,000 images respectively. This shows that the orthogonal matrix with momentum is a better strategy than the random sampling strategy and CMAES.

\begin{figure}[t]
	\centering
	\includegraphics[width=0.36\textwidth]{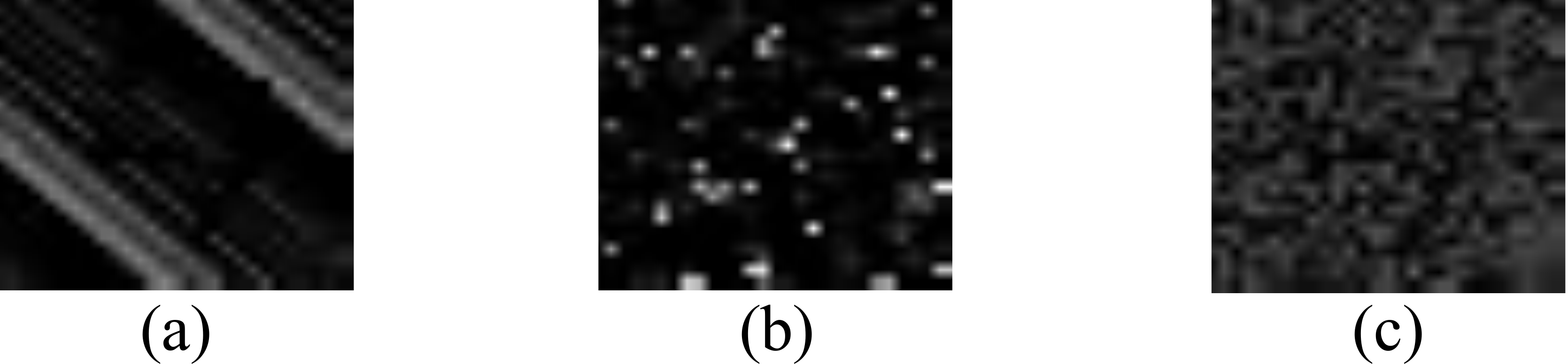}
	\caption{Universal adversarial perturbations computed by DUAttack using different update strategies. (a)(b)(c) represent the universal perturbation computed by DUAttack using the orthogonal matrix with momentum, CMAES~\cite{dong2019efficient} and random sampling respectively.}
	\label{fig_strategy}
\end{figure}

\begin{figure}[t]
	\centering
	\includegraphics[width=0.46\textwidth]{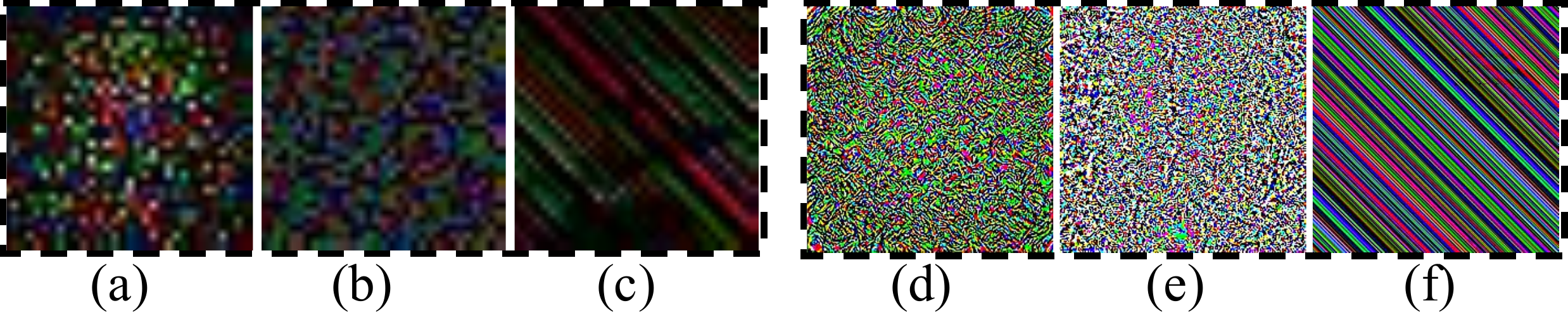}
	\caption{(a)(d),(b)(e),(c)(f) represent the universal adversarial perturbation computed by UAP~\cite{moosavi2017universal}, SingularFool~\cite{khrulkov2018art} and DUAttack respectively. Perturbations on (a)(b)(c), (d)(e)(f) are crafted on CIFAR10 and ImageNet respectively. Pixel values are scaled for visibility.}
	\label{fig_perturbation}
\end{figure}

To further analyze why the orthogonal matrix with momentum strategy can let DUAttack have better performance, we visualize the universal adversarial pertubation $ \mathbf{V} $ crafted by DUAttack using the random sampling strategy, CMAES and orthogonal matrix with momentum strategy as the update strategy respectively. As shown in Fig.~\ref{fig_strategy} and Fig.~\ref{fig_perturbation}, the universal perturbation computed by DUAttack using the orthogonal matrix with momentum strategy has more texture like diagonal stripe. Alex Krizhevsky \etal~\cite{krizhevsky2012imagenet} showed convolutional kernels learned by the first convolutional layer of AlexNet, it shows that CNN especially the top convolutional layers are sensitive to texture like horizontal stripe, barre mark and diagonal stripe. 
We also find such results from the experiments that the proposed method DUAttack with the orthogonal update strategy has better performance than other methods.

Besides, during the updating process, DUAttack ethier plus or minus with the perturbation. We further visualize the distribution of these two operations during the updating process. As shown in Fig.~\ref{fig_plusminus}, in general, the operation plus are more likely to be choosen for computing the perturbation. 
\begin{figure*}[t]
	\centering
	\includegraphics[width=0.93\textwidth]{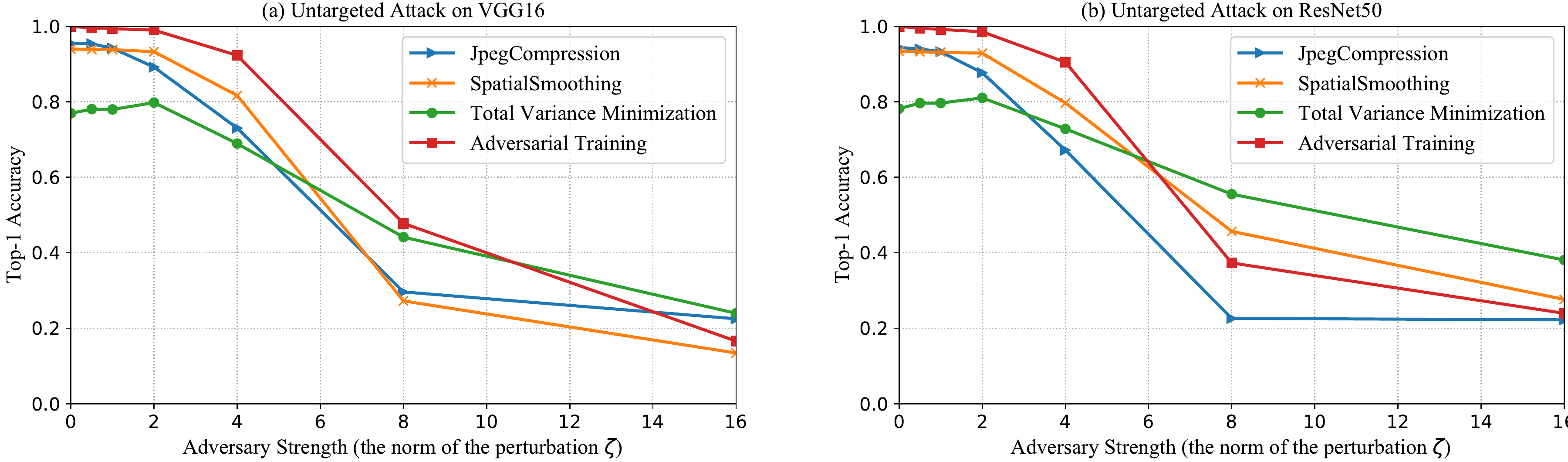}
	\caption{(a) (b) represent the classification accuracy of VGG16 and ResNet50 trained with defenses respectively. DUAttack is applied to these models on CIFAR-10. The perturbation for attacking ResNet50 is generated by VGG16.}
	\label{fig_defend}
\end{figure*}
\subsection{Attack against defenses}
We have investigated the universality of the universal adversarial perturbation computed by the proposed method DUAttack across data, models, and tasks, and have examined the performance of our adversary with different sizes of the training images. Now we evaluate the performance of our attack against defenses.

We choose a local spatial smoothing method Spatial Smoothing (SS)~\cite{xu2017feature}, a compression approach JPEG Compression (JC)~\cite{das2017keeping}, a method modifying the adversarial examples Total Variance Minimization (TVM)~\cite{guo2018countering} and an approach modifying the training schemes Adversarial Training (AT). We train VGG16 and ResNet50 with SS, JC, TVM and AT on CIFAR-10. When applying AT, we add the natural images with the universal perturbation. The perturbations are all computed for VGG16, then they are added to natural images for attacking ResNet50. We perform DUAttack against these defenses using the perturbation with the desired norm ranging from $0$ to $16$. The step size and the number of iteration for DUAttack are 0.2, 1000 respectively. We utilize Adversarial Robustness Toolbox~\cite{art2018} to implement SS, JC, TVM.

As illustrated in Fig.~\ref{fig_defend}, the models have the highest top-1 accuracy when trained using AT. VGG16 and ResNet50 achieve about 100\% accuracy on 10,000 test images of CIFAR-10. On the contrary, models trained with TVM have the lowest accuracy, but can partially eliminate the effect of DUAttack. It should be noted that when the desired $\ell_F$ norm of the universal perturbation is about 8, the accuracy of models with these defenses drops to around 50\%. When the distance of perturbation generated by our method becomes larger ($\zeta = 16$), the accuracy of models is close to $20\%$, the classification ability of models has basically been lost.

Figure~\ref{fig_defend} shows that defenses including modifying the input (gradient masking) and modifying the training schemes cannot always perform well against DUAttack. The adversary only has access to the final labels returned by the model but can attack unseen images and models by adding the universal adversarial perturbation learned from one model, even these unseen models have defense methods.

\subsection{Experiments on real-world settings}
In this subsection, to validate the effectiveness of the proposed method DUAttack against real world platforms, we compare the proposed method with transfer-based attack PGD, UAP and query-based attack including SimBA on the online model from Microsoft Azure. 

\vspace{+0.5em}
\noindent
\textbf{Microsoft Azure}
\begin{table}[t]
	\begin{center}
		\small
		\caption{Fooling rates (\%) on online Azure model on MNIST. The numbers in ($\cdot$) denote the average $\ell_F$ perturbation distance per image. $1^{st} / 2^{nd}$ best in \textbf{bold}/\textit{italic}.} 
		\label{tab_Azure}
		\begin{tabular}{lccc}
			\hline
			&Method  &Untargeted &Targeted \\
			\hline
			\multirow{2}{*}{Individual}  &PGD     &67.41(4.75)  &23.74(5.43) \\
			&SimBA  &\textbf{83.50(1.17)}  &\textbf{46.56(1.69)} \\
			\hline
			\multirow{2}{*}{Universal}  &UAP           &56.24(7.11)  &31.65(5.29) \\
			&DUAttack  &\textit{83.08(4.73)}  &\textit{40.47(4.71)} \\
			\hline
		\end{tabular}
	\end{center}
\end{table}
We conduct experiments on attacking the online model from Microsoft Azure in two scenarios. We use the example MNIST model of the machine learning tutorial on Azure as the attacked model and employ it as a web service. We do not know the machine learning method and architecture of this model. The only information we can obtain is the outputs of this model.

To compare with the performance of other attack methods in black-box setting. We utilize PGD and UAP to generate the perturbation on a 4-layer CNN in white-box setting, and apply the perturbation to the online MNIST model. We apply SimBA and our DUAttack to the online model directly in black-box setting. The results are shown in Table \ref{tab_Azure}. This experiment in the real-world environment validates our hypothesis -- the attacks without transfer obviously outperform the attacks generated from other models (SimBA $>$ PGD, DUAttack $>$ UAP). Even our universal DUAttack performs better than the individual attacks PGD. The results show that our universal perturbation without transfer is significantly better than the perturbation generated by UAP, which needs transfer for achieving black-box attack.

\section{Conclusion}
In this work, we have proposed a method to perform the universal attack which only has access to the top-1 predicted labels returned by the attacked models. The learned universal perturbation can be applied to all benign images on different models and tasks. Experimental results confirmed that the stripe texture adversarial perturbation have high transferability, and the adversary can still work well against several defenses and on the real-world models. 

In the future, we will design adaptive adjustment parameters to find minimal perturbation which is sufficient to fool models. Furthermore, we will study on perturbations having more kinds of patterns, and improving the robustness of models to the universal adversarial perturbation.

\bibliography{aaai21}

\end{document}